\documentclass{sig-alternate} 
\usepackage{mathptmx} 

\newcommand{\ignore}[1]{}
\usepackage{fancyhdr}
\usepackage[normalem]{ulem}
\usepackage[hyphens]{url}
\usepackage{microtype}

\usepackage{hadianeh}
\usepackage{macros}
\usepackage{numbers}
\usepackage{names}

\renewcommand{\baselinestretch}{0.9788}

\fancypagestyle{firstpage}{
}  

\title{\huge{\vspace{-0.45cm}\protect~ORIGAMI: A Heterogeneous Split Architecture for In-Memory Acceleration of Learning\vspace{-0.46cm}}} 
\author{
\begin{tabular}{cccc}
\fontsize{10}{12}\selectfont{}Hajar Falahati$^\S$$^\ddagger$&\fontsize{10}{12}\selectfont{}Pejman Lotfi-Kamran$^\ddagger$&\fontsize{10}{12}\selectfont{}Mohammad Sadrosadati$^\flat$&\fontsize{10}{12}\selectfont{}Hamid Sarbazi-Azad$^\flat$$^\ddagger$
\end{tabular}
\cr
\begin{tabular}{cccc}
\fontsize{9}{16}\selectfont{}$^\S$Iran University of Science and Technology&\fontsize{9}{16}\selectfont{}$^\ddagger$Institute for Research in Fundamental Sciences (IPM)&\fontsize{9}{16}\selectfont{}$^\flat$Sharif University of Technology&\fontsize{9}{16}
\end{tabular}
\cr
\begin{tabular}{ccccc}
\fontsize{9}{16}\selectfont{}$^\S$hfalahati@iust.ac.ir&\fontsize{9}{16}\selectfont{}$^\ddagger$\{hfalahati, plotfi, azad\}@ipm.ir&\fontsize{9}{16}\selectfont{}$^\flat$sadrosadati@ce.sharif.edu&\fontsize{9}{16}\selectfont{}$^\flat$azad@sharif.edu&\fontsize{9}{16}
\end{tabular}  
}

\begin{document}
\maketitle
\thispagestyle{firstpage}
\pagestyle{plain}

\begin{abstract}
One of the major challenges in processing machine learning (ML) algorithms is the memory bandwidth bottleneck. 
In-memory acceleration has the potential to address this problem. However, a solution based on in-memory acceleration needs to address two challenges.
First, in-memory accelerators should be general enough to support a large set of different ML algorithms.
Second, the solution should be efficient enough to utilize the bandwidth while meeting the limited power and area budgets of the logic layer of a 3D-stacked memory.
We observe that previous work fails to simultaneously address both challenges. 

In this work, we propose~\hpim~that includes a heterogeneous set of in-memory accelerators to support compute demands of different ML algorithms, and also uses an off-the-shelf compute platform (e.g., FPGA, GPU, TPU, etc.) in coalescence with the in-memory accelerators to utilize the bandwidth without violating the strict area and power budgets. 
\hpim~offers a pattern-matching technique to identify the similar patterns of computation across a set of ML algorithms and extracts a compute engine for each pattern. 
These compute engines constitute the heterogeneous accelerators that are integrated on the logic layer of a 3D-stacked memory.
The combination of these compute engines can execute any type of ML algorithms.
To utilize the available bandwidth without violating area and power budgets of the logic layer, \hpim~comes with a computation-splitting compiler that divides an ML algorithm between the in-memory accelerators and an out-of-the-memory platform in a balanced way and with minimum inter-communications.

The combination of pattern matching and split execution offers a new design point for the acceleration of ML algorithms.
The evaluation results across 12 popular ML algorithms show that \hpim~outperforms the state-of-the-art accelerator with 3D-stacked memory in terms of performance and energy-delay product (EDP) by \xxs{1.5$\times$} and \xxs{29$\times$} (up to \xxs{1.6$\times$} and \xxs{31$\times$}), respectively.
Furthermore, the results are within a 1\% margin of an ideal system that has unlimited compute resources on the logic layer of a 3D-stacked memory.
\end{abstract}
\section{Introduction}
\label{sec:intro}

Machine learning (ML) is set out to revolutionize the way that the individuals and the society interact with and utilize the machines.
These advances; however, are predicated on delivering high-performance platforms for training models during the training phase. The trained model, then,  is used to evaluate unseen data, a.k.a., the inference phase. 
%
Training ML models is significantly compute intensive and at the same time, puts a lot of pressure on the memory~\cite{eyeriss:isca:2016, shidiannao:isca:2015, imagenet:neural:2012, very:arxiv:2014, cnvlutin:rchnews:2016, tetris:asplos:2017, pipelayer:hpca:2017, graphpim:hpca:2017, tom:isca:2016, isaac:archnews:2016, neurocube:isca:2016, chameleon:micro:2016, prime:archnews:2016, pim:nda:hpca:2015, pim:graph:isca:2015, pim:drama:cal:2015, pim:sparse:hpec:2013, neuflow:cvprw:2011}.
%
Given these characteristics, in-memory acceleration~\cite{isaac:archnews:2016, pipelayer:hpca:2017, prime:archnews:2016, tetris:asplos:2017, graphpim:hpca:2017, tom:isca:2016, neurocube:isca:2016, chameleon:micro:2016, pim:nda:hpca:2015, pim:graph:isca:2015, pim:drama:cal:2015, pim:sparse:hpec:2013, neuflow:cvprw:2011, CMP-PIM_DAC118, PIM-training_Micro18} is a natural fit for accelerating ML algorithms.
%

By advent of 3D-stacked memories~\cite{micro:hmc:spec, MICRON, phd:hmc, soc:dic3:2010, smartrefresh}, in-memory acceleration~\cite{tetris:asplos:2017,graphpim:hpca:2017,tom:isca:2016,neurocube:isca:2016,chameleon:micro:2016,pim:nda:hpca:2015,pim:graph:isca:2015,pim:drama:cal:2015,pim:sparse:hpec:2013,neuflow:cvprw:2011, CMP-PIM_DAC118, PIM-training_Micro18} becomes a feasible solution.    
%
Various pieces of inspiring work have devised in-memory accelerators for ML algorithms but mostly focused on the inference phase~\cite{tetris:asplos:2017, graphpim:hpca:2017, tom:isca:2016, chameleon:micro:2016, pim:nda:hpca:2015, pim:graph:isca:2015, pim:drama:cal:2015, pim:sparse:hpec:2013, neuflow:cvprw:2011, machine:neurocomputing:2017} or training phase of special kinds of ML algorithms~\cite{neurocube:isca:2016} that can be done by exploiting compute units of the inference phase, i.e., multiply-accumulator (MAC) units.
%

An ideal in-memory accelerator for training ML algorithms should be (1) \codebold{general} to support different kinds of ML algorithms, as there are variations in the compute patterns of different ML algorithms, and (2) \codebold{efficient} to capture the available bandwidth of 3D-stacked memories~\cite{phd:hmc, MICRON, micro:hmc:spec}, while meeting the limited power and area budgets of these memories. 
We observe that previous work limits the potential capability of in-memory accelerators as none of them provides all the necessary features (see \S~\ref{sec:motiv} for more detail). As an example, integrating general-purpose units~\cite{tabla:hpca:2016} inside the 3D-stacked memory only captures up to \xx{16\%} of the available bandwidth (see \S~\ref{sec:motiv} for more detail). 

We set out to explore an in-memory acceleration with heterogeneous compute units to support a wide range of ML algorithms.
Investigating a wide range of popular ML algorithms, we observe that ML algorithms exploit common compute patterns, in which, each pattern can be executed on a specialized compute unit with low area and power overheads. 
The combination of these compute units can execute any type of ML algorithms.
Constrained by the limited area and power budgets of a 3D-stacked memory, even these highly optimized compute units capture only \xx{47\%} of the memory bandwidth (see \S~\ref{sec:motiv} for more details).
Although the captured bandwidth is much larger than that of the general-purpose units, it is still lower than the total available bandwidth.
We conclude that in-memory accelerators, alone, cannot utilize the whole available bandwidth even if we use light-weight compute units due to the limited area and power budgets of the 3D-stacked memory.

To capture all the available bandwidth, we aim to enable a split execution between the light-weight heterogeneous in-memory engines and an out-of-the-memory compute platform. We observe that existing 3D interfaces can transfer about 63\% of the internal bandwidth to an out-of-memory compute platform~\cite{MICRON, micro:hmc:spec}.
Inside the memory, we integrate as many compute units as the area and power budgets allow to capture the 3D-stacked memory bandwidth. We drive an out-of-the-memory compute platform by the unused portion of the bandwidth. 
To fully utilize the two platforms, we observe that ML algorithms are composed of many parallel regions, which facilitate execution of ML algorithms over the two platforms, in-memory accelerators and an out-of-the-memory platform, with minimal inter-communications.
%

\hpim\footnote{The small number of basic origami folds can be combined in a variety of ways to make intricate designs.} is a hardware-software solution that combines compute patterns with a heterogeneous set of in-memory accelerators, and splits the execution over the in-memory accelerators and an out-of-the-memory platform.
 \hpim~translates the common compute patterns of different ML algorithms into heterogeneous compute engines that should be integrated on the logic layer of 3D-stacked memories.
Moreover, \hpim~efficiently distributes parts of the computation of an ML algorithm to an out-of-the-memory compute platform to capture all of the available memory bandwidth provided by a 3D-stacked DRAM.

This paper makes the following contributions:
\begin{itemize}[leftmargin=2.5mm,itemsep=0mm,parsep=0mm,topsep=0mm]
\item We extract common compute patterns and parallelism types of a set of different ML algorithms. 
\item We propose \hpim~that benefits from a set of heterogeneous in-memory accelerators derived from the identified compute patterns, and splits the computation over the in-memory accelerators and an out-of-the-memory compute platform using the identified parallelism types. 
\item We show that \hpim~outperforms the state-of-the-art solution in terms of performance and energy-delay product (EDP) by \xxs{1.5$\times$} and \xxs{29$\times$} (up to \xxs{1.6$\times$} and \xxs{31$\times$}), respectively.
Moreover, \hpim~is within a 1\% margin of an ideal system, which has unlimited compute resources on the logic layer of a 3D-stacked memory.
\end{itemize}
\section{Motivation}
\label{sec:motiv}

There are two  phases in processing ML algorithms: (1) a \emph{training phase} that optimizes the model parameters over a training dataset, and (2) an \emph{inference phase} where the trained model is deployed to process new unseen data.
While both phases are computationally intensive, the training phase demands more compute resources due to two reasons. First, the training phase is a superset of the inference phase. The inference phase just includes multiply-accumulator (MAC) operations, but the training phase, which  optimizes different objective functions, includes more operations like non-linear operations. Second, to achieve high accuracy for the trained model, ML algorithms require copious amounts of processing power to iterate over vast amounts of training data~\cite{eyeriss:isca:2016, shidiannao:isca:2015, imagenet:neural:2012, very:arxiv:2014, isaac:archnews:2016, cnvlutin:rchnews:2016}.

Intrinsic parallelism of ML algorithms has inspired both academia and industry to explore accelerating platforms such as FPGAs~\cite{FPGADeep:FPGA:2015, dnnweaver:micro:2016,tabla:hpca:2016, CHiMPS:FPGA:2008, large:journal:2011, cosmic:Micro:2017, resourcePartitioning:ISCA:2017}, GPUs~\cite{convGPU:2017, Oh:2004, Guzhva:2009, saberlda:asplos:2017}, and ASICs~\cite{bitfusion:isca:2018, snapea:isca:2018, ganax:isca:2018, eyeriss:isca:2016, cnvlutin:rchnews:2016, shidiannao:isca:2015, stripes:micro:2016, scnn:isca:2017, eie:isca:2016, cambrion:micro:2016, dadiannao:micro:2014, pudiannao:asplos:2015, tpu:isca:2017, cosmic:Micro:2017, scalpel:ISCA:2017, scaledeep:ISCA:2017}.
%
However, the high memory footprints of ML algorithms limit the potential performance benefits of accelerations.
\subsection{Memory Bandwidth Bottleneck}
To keep compute resources busy, accelerators need to transfer huge amounts of data, which makes memory subsystem a serious bottleneck in terms of bandwidth and energy~\cite{eyeriss:isca:2016, shidiannao:isca:2015, imagenet:neural:2012, very:arxiv:2014, cnvlutin:rchnews:2016, tetris:asplos:2017, pipelayer:hpca:2017, graphpim:hpca:2017, tom:isca:2016, isaac:archnews:2016, neurocube:isca:2016, chameleon:micro:2016, prime:archnews:2016, pim:nda:hpca:2015, pim:graph:isca:2015, pim:drama:cal:2015, pim:sparse:hpec:2013, neuflow:cvprw:2011, tabla:hpca:2016, dnnweaver:micro:2016, machine:neurocomputing:2017, saberlda:asplos:2017}.
A large body of work has explored in-memory processing, built upon DRAM~\cite{dadiannao:micro:2014}, SRAM~\cite{PIM-SRAM:VLSI:2016}, non-volatile memories~\cite{prime:archnews:2016, isaac:archnews:2016, pipelayer:hpca:2017}, and 3D-stacked memories~\cite{tetris:asplos:2017, graphpim:hpca:2017, tom:isca:2016, neurocube:isca:2016, chameleon:micro:2016, pim:nda:hpca:2015, pim:graph:isca:2015, pim:drama:cal:2015, pim:sparse:hpec:2013, neuflow:cvprw:2011, machine:neurocomputing:2017} for performance improvements and energy savings.

Many pieces of prior work proposed in-memory accelerators that are built upon 3D-stacked memories, as these memories are commercialized (e.g., HMC~\cite{MICRON, micro:hmc:spec}) and in-memory processing within them is feasible. 3D-stacked memories stack multiple DRAM dies on top of each other inside a package. These dies are vertically connected via thousands of low-capacitance through-silicon vias (TSVs) to a logic die in which the memory controllers are located. 3D memories use high-speed signaling circuits from the logic die to the active die (e.g., CPU, GPU, FPGA, etc.) out of the memory. Putting all together, 3D-stacked memories provide massive bandwidth with low access energy (3 to 5 times smaller) as compared to the conventional DRAMs.  

\subsection{Challenges of In-memory Acceleration}
While in-memory accelerators have the potential to address the memory bandwidth bottleneck, they have two main challenges that should be addressed. 
First, there are many types of ML algorithms and an in-memory accelerator should be able to effectively accelerate them. 
Second, there is a significant constraint on the area and power usage of the logic die in 3D-stacked  memories~\cite{graphpim:hpca:2017, soc:dic3:2010, pim:isca:2015, pim:graph:isca:2015, machine:neurocomputing:2017, isaac:archnews:2016, neurocube:isca:2016, pipelayer:hpca:2017, prime:archnews:2016}.

To evaluate in-memory accelerators, we define two parameters:
(1) \textit{Generality}: how much the architecture is flexible to support different kinds of ML algorithms; 
(2) \textit{Efficiency}: how much the architecture can utilize the available bandwidth subject to area and power constraints.

To achieve generality in accelerating a wide range of ML algorithms with different objective functions, an in-memory accelerator may
 use general-purpose execution units to execute different operations including various non-linear operations (e.g., Sigmoid) in the training phase.
General-purpose execution units can provide generality. However, they are expensive in terms of area and power.
On the other hand, to achieve efficiency, we need to integrate as many general-purpose execution units as needed to capture the whole available bandwidth.

Prior 3D-stacked based in-memory accelerators either support the inference phase~\cite{tetris:asplos:2017, graphpim:hpca:2017, tom:isca:2016, chameleon:micro:2016, pim:nda:hpca:2015, pim:graph:isca:2015, pim:drama:cal:2015, pim:sparse:hpec:2013, neuflow:cvprw:2011, machine:neurocomputing:2017} or the training phase of special kinds of ML algorithms such as Convolutional Neural Network (CNN)~\cite{neurocube:isca:2016}. 
As there is \textit{no} previous in-memory accelerator that supports the training phase of different types of ML algorithms, we implement general-purpose units similar to those in prior work~\cite{tabla:hpca:2016, cosmic:Micro:2017}. Considering the available power and area budgets of 3D-stacked memories, these general-purpose units only capture \xx{80~GB/s} (\xx{16\%}) of the available bandwidth (out of \newtext{\xx{512~GB/s}}, more details in \S~\ref{sec:eval}).
The captured bandwidth is much lower than the total bandwidth of 3D-stacked memories, which shows that general-purpose in-memory accelerators fail to utilize the available bandwidth.
\subsection{Holistic In-memory Approach}
 
To alleviate the bandwidth bottleneck and accelerate the training phase of a wide range of ML algorithms, we propose a holistic in-memory approach, called \hpim, which satisfies both ML requirements and limitations of 3D-stacked memories. 
While we focus on accelerating the training phase of ML algorithms, the proposed idea can also be applied to the inference phase, as the training phase is a superset of the inference phase.

\hpim~benefits from low-overhead compute engines as in-memory accelerators on the 3D-stacked memory to capture as much 3D-stacked memory bandwidth as possible (240~GB/s of 512~GB/s). 
\hpim~uses the rest of the bandwidth (272~GB/s of 512~GB/s) to drive an out-of-the-memory compute platform, which can be ASIC, GPU, FPGA, TPU, or any other types of compute platform.

We build \hpim~upon two key ideas: 

\niparagraph{Pattern-Aware Execution.}
\hpim~exploits pattern-aware execution to accelerate different ML algorithms by light-overhead heterogeneous compute engines on the logic die of a 3D-stacked memory.
First, \hpim~identifies these compute patterns. Second, \hpim~implements each compute pattern by a specific hardware unit, called compute engine, with low area and power overheads. 
The combination of these heterogeneous compute engines can accelerate different types of ML algorithms. 

\niparagraph{Split Execution.}
3D-stacked memories provide the active die with external bandwidth of up to \xx{320~GB/s}.
This observation motivates us to integrate as many compute engines as possible on the logic die and feed the unused portion of the available bandwidth to the active die.
The combination of accelerators on the logic die and the compute platform on the active die utilizes the whole 3D-stacked DRAM bandwidth.

%
%
%
To partition an ML algorithm efficiently between the in-memory accelerators and the out-of-the-memory platform, a partitioning algorithm should have three features.
(1) \emph{Concurrency:} Partitioned parts should be able to run simultaneously. 
(2) \emph{Minimum inter-communications:} Partitioned parts should have no or minimum inter-platform communications. 
(3) \emph{Load Balancing:} Partitioned parts should be proportional to the compute capabilities of the two platforms.
Compute capabilities are limited by two factors. First, compute throughput that depends on the speed of hardware. Second, memory throughput that depends on the available memory bandwidth.
%
%
To have all of these features, we observe that there are various parallelism types inside ML algorithms. 
\hpim~extracts three parallelism types from ML algorithms, that guarantee concurrency and minimum inter-communications.
Then, \hpim~exploits an assignment algorithm to partition these concurrent parts based on the compute capabilities of the two platforms, that guarantees load-balancing.

We compare some prior work~\cite{tabla:hpca:2016, scalpel:ISCA:2017, scaledeep:ISCA:2017, resourcePartitioning:ISCA:2017, neurocube:isca:2016, PIM-training_Micro18, CMP-PIM_DAC118} in Table~\ref{tab:previous}.
\emph{In-memory} and \emph{Training} columns show if the method uses an in-memory accelerator and targets the training phase, respectively. 
The other five columns show whether the method offers generality, split execution, concurrency, load-balancing, and minimum inter-communications, respectively. 
As summarized in Table~\ref{tab:previous}, pieces of prior work that use in-memory accelerators~\cite{CMP-PIM_DAC118, neurocube:isca:2016, PIM-training_Micro18} do not support execution of the training phase of different kinds of ML algorithms. 
While \emph{TABLA}~\cite{tabla:hpca:2016} is a general method to accelerate the training phase of ML algorithms, it suffers from memory bandwidth problem.
Other pieces of work~\cite{scalpel:ISCA:2017, scaledeep:ISCA:2017, resourcePartitioning:ISCA:2017, PIM-training_Micro18} benefit from split execution but do not support acceleration of the training phase of different kinds of ML algorithms. 
\emph{Scalpel}~\cite{scalpel:ISCA:2017}, \emph{Proger PIM}~\cite{PIM-training_Micro18}, and \emph{Resource partitioning}~\cite{resourcePartitioning:ISCA:2017}) execute only special parts of algorithms over multiple resources and run the rest on just one compute resource. Such techniques fail to provide concurrency and load balancing.
Solutions such as partitioning ML algorithms based on the types of layers, e.g., \emph{Scaledeep}~\cite{scaledeep:ISCA:2017}, which assign memory-intensive parts to the in-memory and compute-intensive parts to the out-of-the-memory platform, neglect the minimum-intercommunications and do not always distribute the computation in a load-balanced manner.
As shown in the table, none of prior work has all the required features.

\begin{table}[!ht]
    \centering
   \caption{Characteristics of previous ML accelerators in supporting: in-memory acceleration (In-memory), training phase (Training), generality (Generality), split execution (Split), concurrency (Conc), load-balancing (L-B), and minimum inter-communications (Min I-C).} 
    \includegraphics[width=0.5\textwidth]{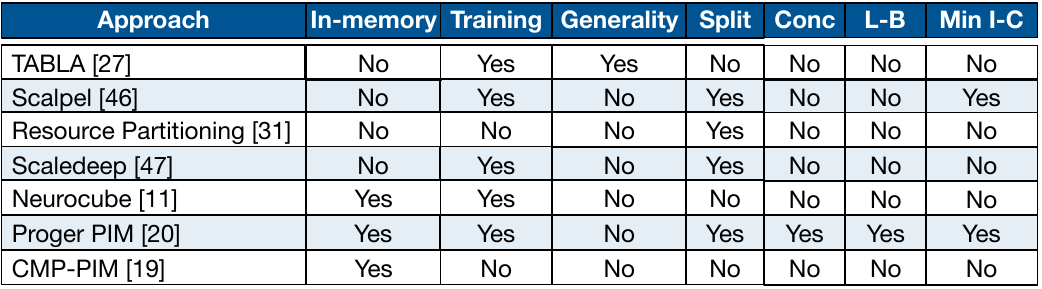}
    \label{tab:previous}
\end{table}
\vspace{-10pt}
\section{Pattern-Aware Execution} 
\label{sec:pattern}

To accelerate different ML algorithms, one solution is to
use general-purpose execution units. However, as we discussed in \S~\ref{sec:motiv}, although general-purpose execution units (e.g.,~\cite{tabla:hpca:2016}) can accelerate a wide range of ML algorithms, they utilize only a small fraction of the available memory bandwidth provided by a 3D-stacked memory due to their large area overhead. 
To address this problem, this work proposes to integrate light-weight accelerators in such a way that the power and area of the accelerators are much smaller that those of general-purpose accelerators, and at the same time, provide enough generality to accelerate different kinds of ML algorithms. Our key idea is to identify common compute patterns of different ML algorithms and map them to light-weight hardware accelerators. 

\subsection{Compute Patterns}
\label{subsec:Blocks}

We thoroughly examine the compute graphs of different ML algorithms and break the graphs down to several compute patterns.
We observe that some of these compute patterns are common across different ML algorithms. 
Each ML algorithm has an objective function, \emph{(f)}, and a set of weights, \emph{(w)}, a.k.a, models,\footnote{We use weight and model interchangeably in this paper.} that map the elements of an input vector, \emph{${(X)}$}, to the output, \emph{${(Y)}$}, as shown in Equation~\ref{eq:objFunc}. 
\begin{equation} \label{eq:objFunc}
\exists W min\sum_{i} f(W, X_{i}, Y_{i})
\end{equation}
The objective function is a cost function that measures the quantity of the distance between the predicted output and the actual output for the corresponding input dataset.
Solving an optimization problem over the training data, ML algorithms minimize the objective function gradually.  

Stochastic Gradient Descent (SGD) is a widely-used algorithm to gradually minimize the objective functions~\cite{de2017understanding, bottou1991stochastic, bottou2012stochastic, asgd, async-p2p, 1bit-sgd, efficient-mini-batch-sgd}. Due to the popularity of SGD, in this paper, we consider SGD as the optimization algorithm, however, our work is general to consider other optimization algorithms as well.

Equation~\ref{eq:sgd} shows how SGD solves the optimization problem defined in Equation~\ref{eq:objFunc}.
\begin{equation} \label{eq:sgd}
W^{(t+1)} =  W^{(t)} - \mu \times \frac{\partial (\sum_{i}f(W^{(t)}, X_{i}, Y_{i})}{\partial (W^{(t)})}
\end{equation}
SGD updates \( W^{(t)} \), by computing \( W^{(t+1)}\), in the reverse direction of the gradient function, \(\partial (f)\), which speeds minimizing the objective function. 
ML algorithms use the updated weights with other $m$ input vectors during the next iterations. 
Parameter \(\mu\) is the learning rate of the ML algorithm. 

Considering the objective functions and parameters of different ML algorithms, we extract four types of compute patterns in the compute graphs of different ML algorithms. The combination of these compute patterns optimizes the objective function.
Three types of these compute patterns are common among different ML algorithms, and one of them is algorithm dependent.
\begin{itemize}[leftmargin=1mm,itemsep=0mm,parsep=0mm,topsep=0mm]
\item  \textbf{Common Compute Patterns.}
\begin{enumerate} [leftmargin=1.5mm,itemsep=0mm,parsep=0mm,topsep=0mm]
	\item \textbf{\macblock~Compute Pattern}. The first compute pattern of different ML algorithms calculates the dot product of the input vector, \emph{(X)}, and the weight vector, \emph{(W)}.  
	We refer to this dot product, \( \sum_{i} X_{i}*W_{{j}{i}} ; i \in [0, k) and  j \in [0, n)\), as \macblock.
	This compute pattern is needed to compute the predicted output.
	\item \textbf{\compblock~Compute Pattern}. The predicted output of an ML algorithm is compared against a threshold or the known output, \emph{(Y)}, usually using a subtractor operation. 
	We refer to this compute pattern as \compblock.
	Using this compute pattern, the output of the objective function, \emph{delta}, is calculated. 
	\item \textbf{\optblock~Compute Pattern}. Using an optimization method, an ML algorithm updates the models to minimize the output of the objective function, \emph{delta}.
	This compute pattern, \(W^{(t+1)} =  W^{(t)} - \mu \times delta \), is referred to as \optblock.
	\end{enumerate}

\item  \textbf{Algorithm-Dependent Compute Pattern.}
In addition to the aforementioned compute patterns, some ML algorithms need to perform extra operations in their objective functions to calculate the predicted output and the delta value.
	These extra operations differ from one ML algorithm to another, and include basic operations \((e.g., -, +, *, <, and >)\) and non-linear operations (e.g., Sigmoid, Gaussian, Sigmoid Symmetric, and Log).
	We refer to this compute pattern as \emph{algorithm-dependent} (a.k.a, \specblock).
\end{itemize}
	%
 %

\begin{table*}[h]
    \centering
   \caption{Compute patterns for several ML algorithms.} 
    \includegraphics[width=0.75\textwidth]{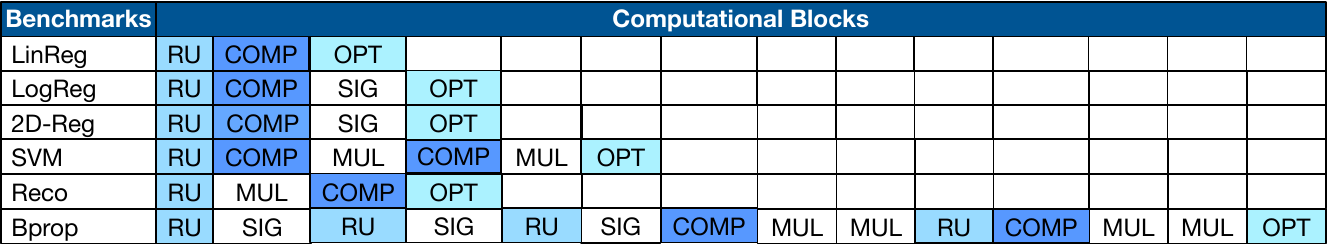}
    \label{tab:pattern}
\end{table*}

\begin{figure}[h]
\setstretch{1}
\centering
\includegraphics[width=0.3\textwidth]{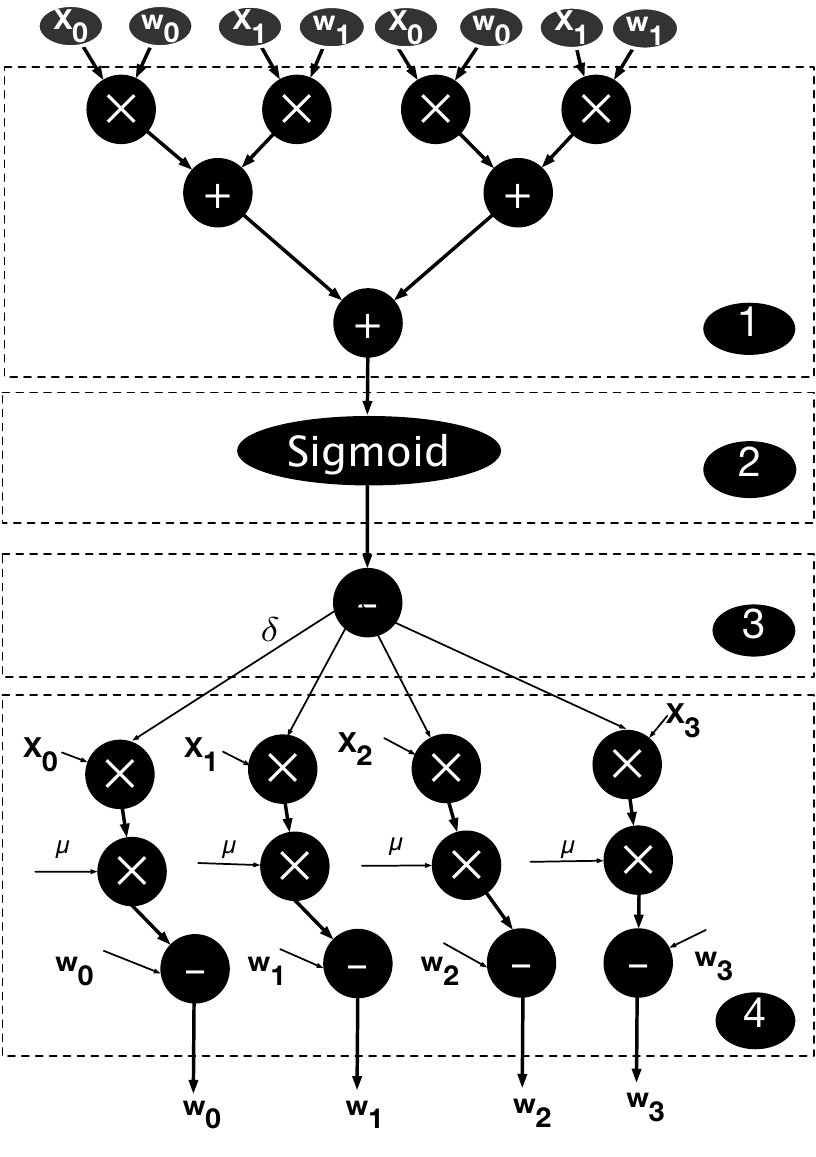}
\caption{Compute patterns of Logistic Regression (LogReg) algorithm.}\label{fig:logistic}
\end{figure}

Table~\ref{tab:pattern} shows the compute patterns of the ML algorithms that we considered.
For example, the compute patterns of \bench{LogReg}, as shown in Figure~\ref{fig:logistic}, include \macblock~(\circledb{1}), \sigblock~(\circledb{2}), \compblock~(\circledb{3}), and \optblock~(\circledb{4}).
This table shows that different ML algorithms have common compute patterns, \macblock, \compblock, and \optblock. Note that \bench{LonReg} has the same compute patterns as \bench{2D-Reg}. The reason is that \bench{LogReg} and \bench{2D-Reg} exploit the same objective function to optimize one-dimensional and two-dimensional models, respectively.

\subsection{Light-Weight Compute Engines} \label{subsec:accelerators}

\begin{figure}[t]
\centering
\includegraphics[width=0.5\textwidth]{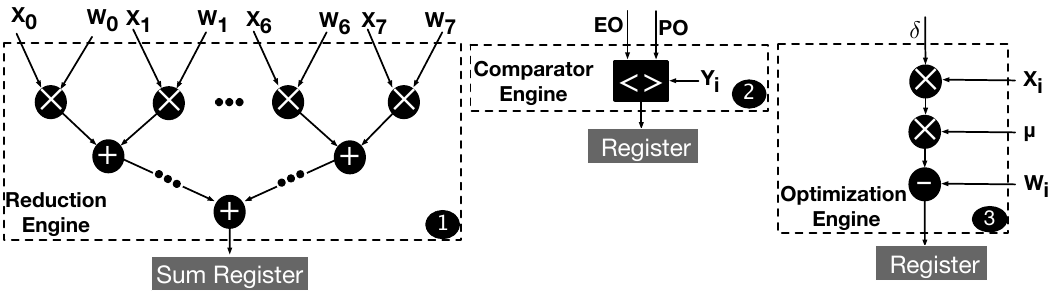}
\caption{Reduction, Comparator, and Optimization compute engines.}\label{fig:arch_detail}
\end{figure}

Compute patterns in the compute graphs of ML algorithms can easily be mapped to a set of \emph{heterogeneous compute engines} such as \macunit, \compunit, \optunit, and \nonlinunit, where the compute engines are customized to execute one particular compute pattern efficiently. 
Each compute engine (accelerator), as shown in Figure~\ref{fig:arch_detail}, is named after its corresponding compute pattern. 

\niparagraph{Reduction Compute Engine (\macunit).}
To perform \macblock~compute pattern, we need an array of multipliers, along with an arrangement of adders to accumulate the products into one final output.
{\macunit} compute engine, $k/RU$, is an array of $k$ multipliers, each evaluating the product of $x_i$ and $w_i$, followed by as many adders as required to aggregate the sum of all the products into one final output, and save the output in the {\emph{sum register}}. 
Figure~\ref{fig:arch_detail}~\circledb{1} shows a \macunit~of size 8, '8/RU', in which eight multiplications are performed, and the products are accumulated using several levels of adders.

\niparagraph{Comparator Compute Engine (\compunit).}
\compunit, labeled as \circledb{2} in Figure~\ref{fig:arch_detail}, performs a subtraction between the predicted output, \emph{PO}, and the expected output, \emph{EO}, and generates a difference to be used for updating the models.

\niparagraph{Optimization Compute Engine (\optunit).}
\optunit, labeled as \circledb{3} in Figure~\ref{fig:arch_detail}, is a serial chain of two multiplier units and a subtractor unit, which updates an element of the \emph{model} array.
To update $n$ \emph{weight}s, we need to integrate $n$ instances of the {\optunit}s.

\niparagraph{Non-Linearity Compute Engine (\nonlinunit).}
The last compute pattern, \specblock, is algorithm-specific and mostly performs non-linear operations. 
Linear operations of \specblock~are performed by other compute engines. 
Regarding the non-linear operations, one realization of this accelerator is a general-purpose ALU that has a special unit for non-linear operations. However, this realization is expensive in terms of area and power. 

To address this problem, we borrow the idea of implementing non-linear functions using lookup tables from prior work~\cite{kaufmann1998signal, engel2001high, keahey1996techniques, mielikainen2006lossless} but instead of lookup tables, we use the die-stacked DRAM to hold the outputs of non-linear functions. This is mainly because the area and power budgets of the logic die are limited and we prefer to dedicate the available area and power to the other compute engines.

\section{Split Execution} 
\label{sec:split}

As the area and power budgets of the logic layer of 3D-stacked memories are limited, in-memory accelerators cannot utilize the whole bandwidth offered by 3D-stacked DRAM even if we use light-weight compute engines.
One way to address this limitation is to use out-of-the-memory resources on the active die in parallel with the in-memory accelerators. The out-of-the-memory resource can be ASIC, GPU, FPGA, TPU, or any other types of compute platforms. 
To this end, we need to split the execution on two platforms. Moreover, split execution offers performance improvement if both platforms are fully utilized and inter-platform communications are minimal.

Analyzing the compute graphs of ML algorithms, we set out a platform-aware partitioning mechanism. 
We observe that there are three different types of parallelism inside ML algorithms. Based on these parallelism types and the specifications of the two platforms (heterogeneous accelerators on the logic die and the out-of-the-memory compute platform), we partition the compute graph over the platforms to maximize resource utilization with minimal inter-platform communications.
	\begin{figure}[t]
	\centering
	\includegraphics[width=0.48\textwidth]{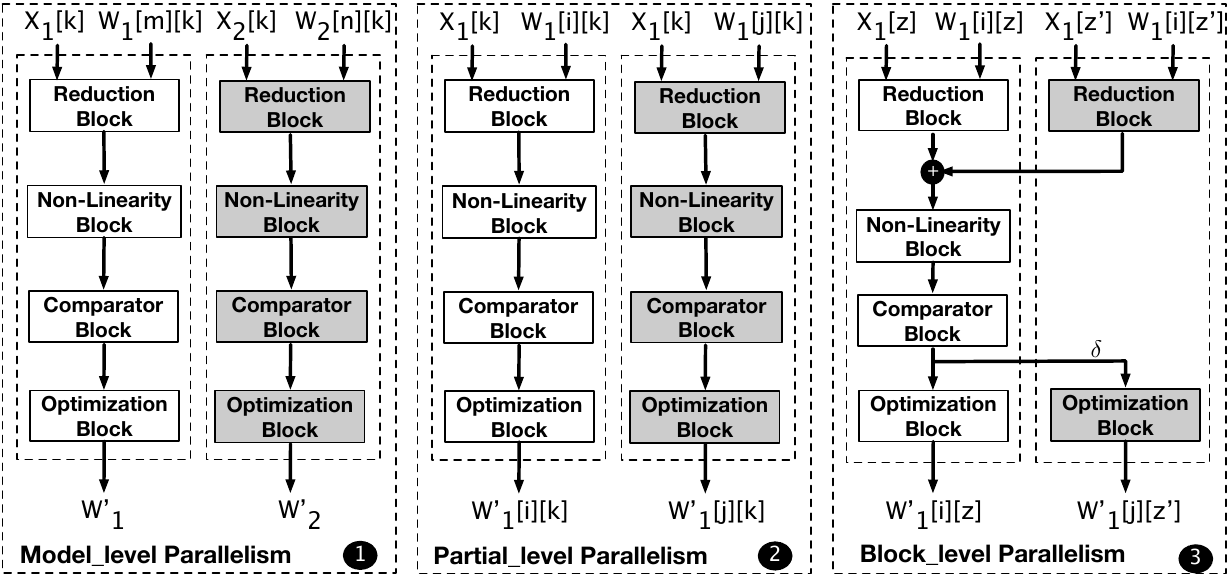}
	\caption{Three different types of parallelism, namely \emph{bloc\_{level}}, \emph{partial\_{level}}, and \emph{model\_{level}}.}\label{fig:parallel}
	\end{figure}
	\vspace{-5pt}
\subsection{Parallelism Types}\label{subsec:parallel}
	We observe that there are up to three types of parallelism, as shown in Figure~\ref{fig:parallel}, in the compute graph of an ML algorithm. 
        \begin{enumerate} [leftmargin=2mm,itemsep=0mm,parsep=0mm,topsep=0mm]
	\item \textbf{Model\_{level} parallelism}. Some ML algorithms optimize more than one \emph{{weight}} array and each \emph{{weight}} array works on completely distinct data arrays. 
	For example, Recommender Systems (\bench{Reco}) algorithm optimizes two independent weights: movie\_{feature} and users\_{feature}. 
	Thus, its compute graph consists of independent compute subgraphs, labeled \circledb{1}~in Figure~\ref{fig:parallel}. 
	\item \textbf{Partial\_{level} parallelism}. The \emph{{weight}} arrays of some ML algorithms such as Recommender Systems (\bench{Reco}), 2D-Regression (\bench{2D-Reg}), and Back-propagation (\bench{BProp}) have more than one dimension, \( W_{{j}{i}};\) \(i \in [0, k)\) and  \(j \in [0, n)\)  where  \($n>1$\). 
	In such cases, the compute graph consists of up to ${n}$ independent compute subgraphs.
	These compute subgraphs work on all elements of an \emph{input} array and a portion of the elements of the \emph{weight} array as labeled 
	\circledb{2}~in Figure~\ref{fig:parallel}. 
	With this type of parallelism, we can create two subgraphs that work on two distinct portions of the {weight} array, \( \emph{model[a][i]} \) and \( \emph{model[b][i]} \) where \( a \in [0, n_1) \), \( b \in [n_1, n) \), and \(i \in [0, k)  \).
	\item \textbf{Block\_{level} parallelism}. There is an internal parallelism in the \macblock~and \optblock~compute patterns as they operate on the \emph{weight} and \emph{input} arrays, \( W_{i} \) and \( X_{i} \) where \( i \in [0, k) \).
	As shown by the compute subgraphs labeled \circledb{3}~in Figure~\ref{fig:parallel}, these compute patterns can be broken into two parts, which operate on distinct elements on \emph{input} and \emph{weight} arrays. 
	While the two parts in the \optblock~compute pattern are independent of each other, the two parts of the \macblock~compute pattern generate partial sums that need to be aggregated (i.e., inter-platform communications). 
	To benefit from this source of parallelism, we partition the {input} and {weight} arrays into two parts. The \macblock~compute pattern can be executed on both partitions in parallel on the two platforms. When the two partitions are executed, the two results are aggregated. To minimize the inter-platform communications, the \optblock~computation will be executed on the same two partitions at each platform.
	\end{enumerate}
	As ML algorithms always include \macblock~and \optblock~compute patterns, our analysis reveals that there is at least one source of parallelism in an ML algorithm (while some algorithms benefit from two or even all three sources of parallelism).
\subsection{Platform-Aware Partitioning}\label{subsec:partition}
Using the three types of  parallelism, we partition an ML algorithm using Algorithm~\ref{alg:patternAlg}. The partitioning algorithm receives the compute graph,~\emph{G}, and the specifications of the two platforms (\codebold{in-memory} and \codebold{out-of-the-memory}, labeled as \codebold{MEM} and \codebold{External} in the algorithm, respectively), and statically partitions the compute graph into two subgraphs to be assigned to the two platforms. 
The partitioning algorithm attempts to maximize the resource utilization using two key ideas: (1) minimizing inter-platform communications, and (2) splitting the execution over the two platforms in a load-balanced manner.
As a result, the algorithm makes sure that the two platforms finish their execution at about the same time.
The static nature of the partitioning algorithm relaxes the hardware control unit from the overhead of runtime load balancing.
%
The \emph{platform-aware partitioning} algorithm follows two steps:
\begin{itemize} [leftmargin=3mm,itemsep=0mm,parsep=0mm,topsep=0mm]
\item \textbf{Minimizing inter-platform communications.} First, we extract all the available types of parallelism in an ML algorithm.
Second, out of the available types of parallelism, we pick the best one to partition the compute graph into two subgraphs (line 5). 
The \emph{model\_{level}} parallelism has the highest priority as it includes two independent subgraphs working on different weights and inputs. 
The next priority belongs to the \emph{partial\_{level}} parallelism in which the compute graph is divided into two independent subgraphs working on distinct parts of the weight array. 
The \emph{partial\_{level}} parallelism has lower priority than the \emph{model\_{level}} parallelism. Unlike \emph{model\_{level}} parallelism, in \emph{partial\_{level}} parallelism, the two partitions operate on the same input.
 The \emph{block\_{level}} parallelism has the lowest priority because it requires inter-platform communications, while the other two types of parallelism (i.e., \emph{model\_{level}} and \emph{partial\_{level}}) have no inter-platform communications. 
\item \textbf{Providing load-balanced partitioning.} To offer a load-balanced partitioning between the two compute platforms, we set the size of each partition (subgraph) based on the throughput of the corresponding platform (lines 7-26). 
To this end, we calculate the throughput of a platform using Equation~\ref{eq:throughput}:
\begin{equation} \label{eq:throughput}
Throughput = min(Memory\,\,BW, \\ Compute\,\,BW)
\end{equation}
For memory bandwidth, we assign as much bandwidth as needed to the heterogeneous compute engines on the logic die. The rest of the bandwidth is given to the out-of-the-memory platform.
To calculate the compute bandwidth, we use Equation~\ref{eq:computeBW}:
\begin{equation} \label{eq:computeBW}
Compute\,\,BW = \sum_{i} \times \sum_{j} I_{{i}{j}} \times data\,\,size \times frequency 
\end{equation}
where \(I_{{i}{j}}\) indicates the number of inputs of the \emph{i}$_{th}$ resource, \emph{data size} indicates the size of each input in byte, and \emph{frequency} is the operating frequency of the platform (lines 7-9). 
Assume that a platform has a $1~GHz$ frequency, $320~GB/s$ \emph{memoryBW}, and 1000 32-bit multipliers. 
Its \emph{compute bandwidth} is \emph{8~TB/s (1000 $\times$ 2 $\times$ 4 $\times$ 1)} and its \emph{throughput} is 320~GB/s. 

To offer a load-balanced partitioning, we partition the compute graph of an ML algorithm based on the throughput ratio of the two platforms, ${rateThroughput}$.
\begin{enumerate} [leftmargin=2mm,itemsep=0mm,parsep=0mm,topsep=0mm]
\item{\textbf{Block\_{level} parallelism.}}
We divide $k$ weights into two parts with proportion to $rateThroughput$, \emph{(k = num1+num2; num1 = rateThroughput $\times$ num2)}, (lines 12-15).
\item{\textbf{Partial\_{level} parallelism.}}
We divide the second dimension of the \emph{weight} array, which its size is $n$, into two parts with the size of $n_1$ and $n_2$ in proportion to ${rateThroughput}$, \emph{( $n$ = $n_1$ $+$ $n_2$; $n_1$ $=$ $rateThroughput$ $\times$ $n_2$)} (lines 16-19).
\item{\textbf{Model\_{level} parallelism.}}
We sort the independent models based on their size. Starting from the largest, we assign models, one by one, to the platform with larger throughput till partitioning ratio becomes $rateThroughput$. As with the  \emph{model\_{level}} parallelism, we only have coarse-grained partitioning ability (i.e., assign the whole model to a platform), the ratio of partitioning might not exactly become $rateThroughput$. Consequently, after \emph{model\_{level}} partitioning is done, we apply other types of available parallelism (e.g.,  \emph{block\_{level}} or \emph{partial\_{level}}) to change the ratio to $rateThroughput$.

\end{enumerate}
\end{itemize} 
\begin{algorithm}[htp]
\setstretch{0.80}
\small
\SetAlgoLined\DontPrintSemicolon
\SetKwFunction{algo}{Platform-Aware Partitioning}\SetKwFunction{proc}{pattern}
\SetKwProg{myalg}{Algorithm}{}{}
\myalg{\algo{}}{
\SetKwInOut{Input}{input}
\SetKwInOut{Output}{output}
  \Input{ 
	\hspace*{0.1cm}$G$: Compute Graph \\
	\hspace*{0.1cm}$External$-$Spec$: Out-of-the-memory Specifications \\
	\hspace*{0.1cm}$MEM$-$Spec$: In-memory Specifications \\
        }
  \Output{
	\hspace*{0.1cm}$External$-$Partitions$: subgraphs assigned to the out-of-the-memory platform \\
	\hspace*{0.1cm}$MEM$-$Partitions$: subgraphs assigned to in-memory compute engines \\
	}

        External-Partitions $\leftarrow$ empty() 	\\
        MEM-Partitions      $\leftarrow$ empty() 	\\
        queue    	    $\leftarrow$ empty() 	\\
         
        pMode = parallelismAnalayzer(G) \\
        queue.push(G)  \\
          
 	External-Spec.throughput = min ( External-Spec.memoryBW, External-Spec.computeBW) \\
        MEM-Spec.throughput   = min (MEM-Spec.memoryBW, MEM-Spec.computeBW) \\
        rateThroughput = External-Spec.throughput / MEM-Spec.throughput \\
        \While {(!queue.empty())}{
                G = queue.pop() \\
		\If{(pMode == Block\_{level})}{
			num1  $\leftarrow$ num-model * (rateThroughput/(1+rateThroughput))\\
			num2  $\leftarrow$ num-model - num1 \\
			{G1, G2} += partition(G, num1, num2) \\
		}
		\ElseIf{ (pMode == Partition\_{level}) }{
			dim1  $\leftarrow$ ${N\_{dim}}$ * (rateThroughput/(1+rateThroughput))\\
			dim2  $\leftarrow$ ${N\_{dim}}$ - dim1\\
                        {G1, G2} += partition(G, dim1, dim2) \\
		}
		\ElseIf{(pMode == Model\_{level})}{
			{g1, g2} $\leftarrow$ partition(G, rateThroughput) \\
			G1 += g1 \\
			rateThroughput = update(g1, g2, rateThroughput)\\
			pMode = parallelismAnalayzer(g2) \\
			queue.push(g2) \\
		} 
	}
	External-Partitions.insert(G1)\\
	MEM-Partitions.insert(G2)
  }{}
  \caption{Platform-Aware Partitioning}\label{alg:patternAlg}
\end{algorithm} 
\section{Origami}
\label{sec:workflow}

\begin{figure*}[ht]
\centering
\includegraphics[width=1\linewidth]{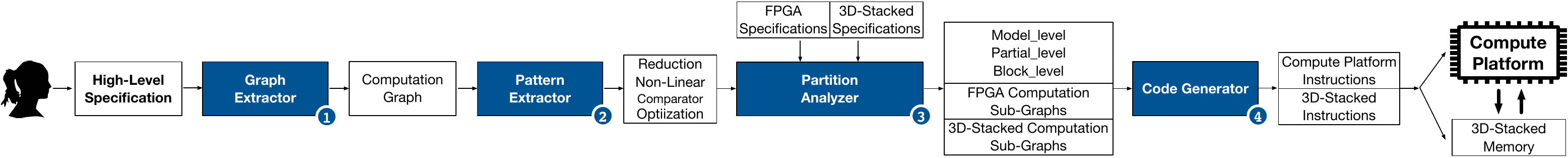}
\caption{{\hpim}~Workflow.}\label{fig:platform}
\end{figure*}
\vspace{-5pt}

%
We propose a heterogeneous split architecture for in-memory acceleration of ML algorithms, called \hpim.
\hpim~is a holistic approach that benefits from pattern and split executions to accelerate different ML algorithms over a set of heterogeneous compute engines on the logic die and a compute platform on the active die.
\hpim~is a hardware-software solution that spreads at different abstraction levels including \emph{programming layer} (\S ~\ref{subsec:PL}), \emph{compiler layer} (\S ~\ref{subsec:CL}), \emph{architecture layer} (\S ~\ref{subsec:AL}), and \emph{hardware layer} (\S ~\ref{subsec:HL}). Figure~\ref{fig:platform} shows the main components of \hpim.
\\
\subsection{Programming Layer}\label{subsec:PL}
		%
		The \emph{programming layer} includes a \emph{programming interface} and a \emph{graph extractor} unit to translate the high-level specification of an ML algorithm to its corresponding \emph{compute graph}, as shown in Figure~\ref{fig:platform}, labeled \circledb{1}.  

		\textbf{Programming Interface.} \emph{Programming interface} receives a high-level specification which includes learning parameters, data declaration, and mathematical declaration of an ML algorithm.
		%
		The learning parameters include the learning rate and the number of features. 
		The data declaration specifies various types of data such as the training input vectors (a.k.a., \emph{input} or \emph{model\_input}), the real outputs (a.k.a., \emph{model\_output}), and the weights (a.k.a., \emph{model\_parameters} or \emph{model}). 
		The mathematical declaration specifies how the objective function of the ML algorithm is computed, using mathematical operations, to update the \emph{weights}. 
		The mathematical operations can be expressed in three categories: (1) \emph{basic operations}, such as \(-, +, *, <, >\), (2) \emph{group operations}, such as \(\sum\), \(\ \lVert \rVert \), \(\Pi\), and (3) \emph{non-linear operations}, such as \(Sigmoid\), \(Gaussian\), \(Sigmoid~Symmetric\), and \(Log\). 

		\textbf{Graph Extractor.} By receiving the high-level specification of an ML algorithm, the \emph{graph extractor} unit extracts the corresponding compute graph.
\subsection{Compiler Layer}\label{subsec:CL}
		At the \emph{compiler layer}, \hpim~performs five operations: (1) extracts compute patterns, (2) detects parallelism types, (3) partitions the compute graph into two load-balanced parts with minimum inter-part communications, (4) assigns each part to a compute platform, and (5) schedules the execution of the algorithm over the two platforms.
		Managing these goals at the compiler layer alleviates the runtime overhead and facilitates management of the simultaneous execution, which simplifies the control mechanism in the 3D-stacked memory. 
		 \emph{Compiler layer} consists of two key components, as shown in Figure~\ref{fig:platform}: 

	 	\circledb{2}~\codebold{Pattern extractor} passes three steps: (1) it creates three pattern subgraphs for the three \emph{common compute patterns}, 
		(2) it runs a pattern-matching algorithm, adopted from graph algorithms~\cite{graph-partitioning:kit:2016, graph-clustering:mathematics:2004}, to find all instances of these compute patterns in the compute graph. 
		The remaining parts of the compute graph are instances of the \emph{\specblock~compute pattern},
		and, (3) it clusters all nodes in each instance as a coarse-grained node in the pattern compute graph.

	\circledb{3}~\codebold{Partition analyzer} detects the parallelism types in the pattern compute graph of an ML algorithm and partitions the graph into two parts to be executed on the heterogeneous compute engines on the logic die and an out-of-the-memory compute platform (i.e., FPGA in this paper\footnote{\hpim~is general and can benefit from different kinds of out-of-the-memory compute platforms such as FPGA, GPU, and etc. Without loss of generality, in this work, we assume that the compute platform on the active die is an FPGA. We use FPGA as an example and leave examination of other platforms for the future work.}). 
		\emph{Partition analyzer} follows Algorithm~\ref{alg:patternAlg} to split the pattern compute graph into two load-balanced partitions with minimum inter-platform communications. 

\subsection{Architecture Layer}\label{subsec:AL}
	The \emph{code generator} uses our proposed \emph{instruction set architecture (ISA)} to prepare executable code and static scheduling for the heterogeneous compute engines.
\subsubsection{ISA}\label{subsubsec:ISA} 
The proposed ISA is a RISC instruction set that consists of two flags, two types of registers, \emph{computation} and {synchronization}, and three types of instructions, \emph{communication}, \emph{computation}, and \emph{synchronization}.
The input and output of each compute engine is hardwired to a dedicated \emph{computation} register. In addition to \emph{computation} registers, there are two registers for synchronizing the execution of the compute engines and the out-of-the-memory platform.

\textbf{Communication Instructions} transfer data from memory locations to registers and vice versa (\codebold{mov \%src, \%des}).

\textbf{Computation Instructions} use the compute engines to perform computation in the 3D-stacked memory.
%
There are three computation instructions: 
\begin{enumerate}[leftmargin=1.5mm,itemsep=0mm,parsep=0mm,topsep=0mm]	%

	\item \codebold{reduce \%Num}:
	enables the \codebold{Num}$_{th}$ \macblock~compute engine to operate on its input registers and store the results in the output register.
	\item \codebold{comparator \%Num}: enables the \codebold{Num}$_{th}$ \compblock~to operate on its input registers and store the results in the output register.
	\item \codebold{optimization \%Num}: enables the \codebold{Num}$_{th}$ \optblock~compute engine to operate on its input registers and store the results in the output register.
\end{enumerate}

\textbf{Synchronization Instructions.}
Synchronization instructions handle the required interactions between the two compute platforms. 

\hpim~utilizes three parallelism types to split the execution of the compute graph over the two platforms. 
With \emph{model\_level} and \emph{partial\_level} parallelism types, there is no inter-communications between the two compute platforms, hence, there is no need for synchronization. 
However, with \emph{block\_level} parallelism type, as shown in Figure~\ref{fig:parallel}, there is a need for inter-communications between the two platforms.
In \emph{block\_level}, the heterogeneous compute engines on the logic die execute a part of the reduction and optimization compute pattern, while the out-of-the-memory compute platform executes the rest. 
With the reduction compute pattern, the partial sum of the two platforms need to be aggregated. The optimization compute engine cannot start the execution until the two partial sums are aggregated. 
For this purpose, one platform (called~\emph{master}) is in charge of aggregating the partial sums and generating the final result, while the other platform (called \emph{slave}) should transfer its partial sum to the \emph{master}.
When \emph{master} is done with its partial sum, it needs to wait to receive the partial sum of the \emph{slave}. Receiving the partial sum, the \emph{master} aggregates the partial sums and sends the final result to the \emph{slave}, which is waiting for the result to start execution of the optimization compute pattern.

To this end, the proposed \emph{ISA} uses two flags, \codebold{M\_{ready}} and \codebold{S\_{ready}}, two synchronization registers, \codebold{M\_{delta}} and \codebold{S\_{psum}}, and three instructions, \codebold{check}, \codebold{set}, and \codebold{wait}. 
Without loss of generality, we assume that the out-of-the-memory compute platform is the \emph{master}.
\begin{enumerate}[leftmargin=1.5mm,itemsep=0mm,parsep=0mm,topsep=0mm]

		\item \codebold{set \%f}: sets the value of flag \codebold{\%f}.\\
		After preparing the partial sum, the in-memory controller should write the partial sum to \codebold{S\_{psum}} and set \codebold{S\_{ready}} by the \codebold{set} instruction. 
		The \emph{master} should check the \codebold{S\_{ready}} flag and reads the partial sum from \codebold{S\_{psum}} when the flag indicates it is ready.

    	        \item \codebold{wait \%f}: waits for the flag \codebold{\%f} to set.\\
		The in-memory controller should wait for the \codebold{M\_{ready}} to set and then read the value of \codebold{M\_{delta}} register. 
		The \emph{master} computes the \textit{delta}, which is needed for the optimization compute pattern, writes it to \codebold{M\_{delta}} register, and sets the \codebold{M\_{ready}} flag.
	       	\item \codebold{clr \%f}: resets the value of flag \codebold{\%f}.\\
\end{enumerate}
\vspace{-5pt}
\subsubsection{Static Scheduling} 
\emph{Code generator} unit receives the part of the compute graph that needs to be executed on the heterogeneous compute engines on the logic die, and transforms it into a sequence of instructions to be executed by the in-memory controller, as we explain in \S \ref{subsec:HL}.
	
\subsection{Hardware Layer}\label{subsec:HL}
  \hpim~adds a set of \emph{heterogeneous compute engines} and an \emph{in-memory controller} to the logic die of a 3D-stacked memory.
		\emph{In-memory controller} is a light-weight unit that executes the instructions of \S \ref{subsubsec:ISA}.

\section{Evaluation}
\label{sec:eval}
\subsection{Experimental Setup}
\label{subsec:exp}
%
\begin{table*}[!ht]
    \centering
   \caption{Benchmarks.} 
    \includegraphics[width=0.85\textwidth]{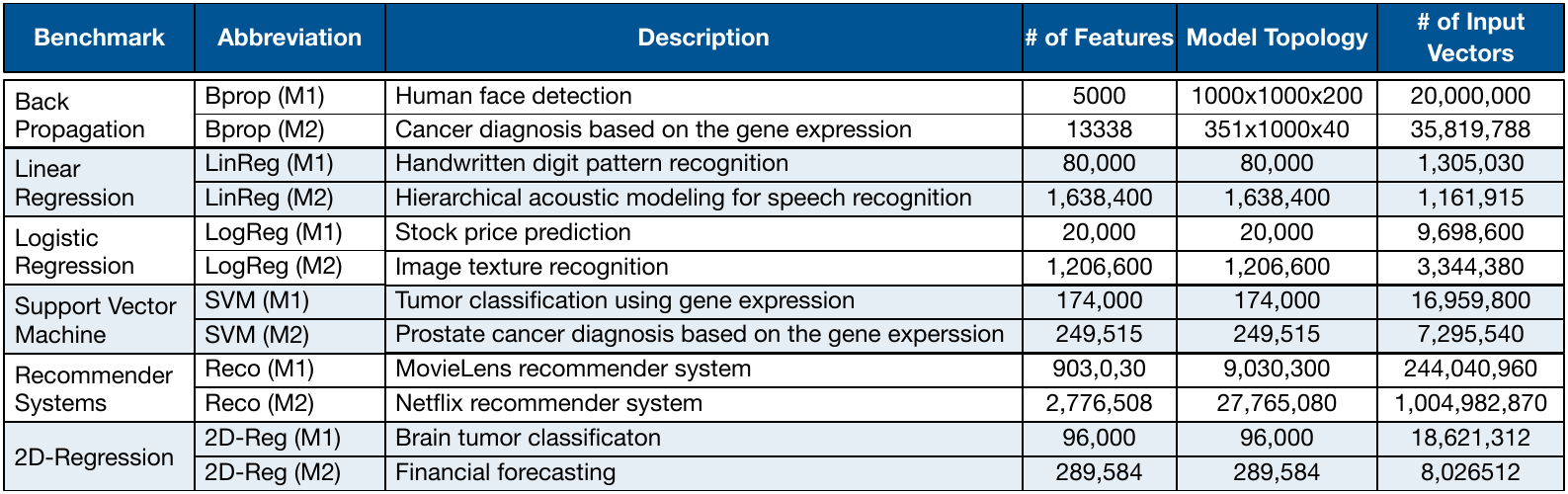}
    \label{tab:bench}
\end{table*}

\niparagraph{Benchmarks and Datasets.}
Table~\ref{tab:bench} summarizes the set of benchmarks that are used for evaluation of \hpim, and their descriptions including model topology, number of features, and number of input vectors.
To evaluate the sensitivity of \hpim's performance improvement to the size of the model used by a given ML algorithm, we use two distinct models (shown as \bench{M1} and \bench{M2} in Table~\ref{tab:bench}) for each evaluated benchmark.
The benchmarks include the state-of-the-art ML algorithms. 
The Back-propagation (\bench{BProp}) algorithm trains models to detect handwritten digits ~\cite{mnist:2010, mnist8m} and speech ~\cite{acoustic:2010}.
The Linear Regression (\bench{LinReg}) algorithm is widely used in finance and image processing to predict prices~\cite{stock-exchange:2008} and texture of images~\cite{texture1:2016}.
The Logistic Regression (\bench{LogReg}) algorithm trains models to detect tumors~\cite{tumor:2003}, and cancer~\cite{cancer1:2002}.
The Support Vector Machine (\bench{SVM}) algorithm is used in computer vision and medical diagnosis domains to detect human faces~\cite{face:2000} and cancer~\cite{cancer2}.
The Recommender Systems (\bench{Reco}) algorithm is widely used in processing movie datasets such as \emph{Movielens} datasets~\cite{movielens:hetrec:2011, movielens_web:2017} and the \emph{Netflix Prize} datasets~\cite{netflix}. 
The 2D-Regression (\bench{2D-Reg}) algorithm trains models to detect different kinds of tumors~\cite{tumor2} and cancers~\cite{cancer2}. 

\niparagraph{FPGA Platform.}
We evaluate \hpim~in the context of a 3D-stacked memory on top of an active die that includes a \code{Virtex UltraScale+ (DS923) VU13P} FPGA.
Table~\ref{tab:arch} reports the key FPGA parameters. 
We synthesize the hardware in the FPGA platform with \code{Vivado Design Suite v2017.2} to extract the FPGA design parameters. 

\niparagraph{ASIC Implementation.}
We use \code{Synopsys Design Compiler (L-2016.03-SP5)} and \code{TSMC\,45-nm} standard cell library at \xx{313} MHz frequency, the frequency of HMC stacked memory~\cite{hybridcube:vlsit:2012,neurocube:isca:2016,micro:hmc:spec,MICRON}, to synthesize the accelerators and obtain the area, delay, and energy numbers.
We use \code{CACTI-P}~\cite{cactip} to measure the area and power of the registers and on-chip SRAMs. 

\niparagraph{Memory Model.}
The 3D-stacked memory is modeled after an HMC stacked memory~\cite{hybridcube:vlsit:2012,neurocube:isca:2016,micro:hmc:spec,MICRON}.
Each vault delivers up to 16 GB/s bandwidth to the logic die and 10 GB/s bandwidth to the active die~\cite{micro:hmc:spec,MICRON}.
The available area to accelerators in each vault is \code{1.5\,mm\textsuperscript{2}}\cite{micro:hmc:spec, tetris:asplos:2017}.
We extract the 3D-stacked memory model parameters from the data sheet~\cite{micro:hmc:spec}.   
Table~\ref{tab:arch} reports the parameters of the memory model used in our evaluations.

\niparagraph{Cycle-Level Simulation.}
Using the ASIC and FPGA synthesis numbers and the configurations of the memory models, we develop a cycle-level architectural simulator to measure the performance and energy consumption of \hpim.
The \hpim~simulator includes the timing of the memory accesses and faithfully models the parameters of the ASIC and FPGA implementations.
Table~\ref{tab:origami} lists the major micro-architectural parameters of \hpim. 
%
\begin{table}[h]
    \centering
    \caption{Major parameters of FPGA and 3D-stacked Memory~\cite{neurocube:isca:2016}.} 
    \includegraphics[width=0.5\textwidth]{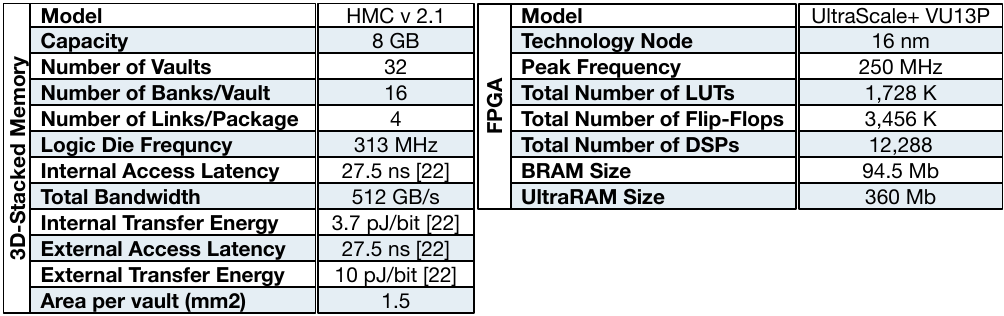}
    \label{tab:arch}
     \vspace{-8pt}
\end{table}
\begin{table}[h]
    \centering
   \caption{Parameters of \hpim.} 
    \includegraphics[width=0.28\textwidth]{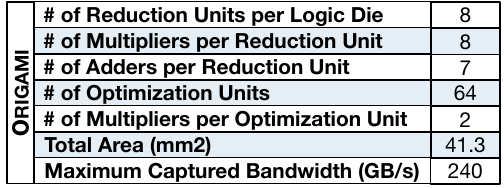}
    \label{tab:origami}
     \vspace{-5pt}
\end{table}
%

\niparagraph{Comparison Metrics.}
We evaluate benefits of \hpim~with six ML algorithms, listed in Table~\ref{tab:bench}, in terms of performance and energy-delay product (EDP).

\niparagraph{Comparison Points.}
We compare six different platforms, namely (1) \codebold{FPGA}, (2) \codebold{PIM-GU}, (3) \codebold{PIM-CE}, (4) \codebold{\hpim}~(our approach), (5) \codebold{\hpim-IIC}, and (6) \codebold{PIM-GU-Unlimited}.

\codebold{\hpim}~represents our approach, in which, we use both the FPGA and the compute engines on the logic die of the 3D-stacked memory.
\hpim~assigns as much of the internal bandwidth as possible to the compute engines and delivers the rest of the bandwidth to the FPGA (See Table~\ref{tab:arch}). Table~\ref{tab:origami} also lists the available resources on the logic die that \hpim~uses for computation.

\codebold{\hpim-IIC} evaluates an \hpim~which exploits an ideal inter-platform communications with no delay and bandwidth usage.
The \codebold{\hpim-IIC} exploits the same configuration as \codebold{\hpim}. We use this comparison point to evaluate the effect of inter-platform communications' delay and bandwidth usage on \hpim's effectiveness.

The state-of-the-art FPGA-based accelerator to train different ML algorithms is \emph{TABLA}~\cite{tabla:hpca:2016}. It has been shown that it outperforms GPU and CPU implementations~\cite{tabla:hpca:2016}.
We implement ALUs of \emph{TABLA} in an FPGA connected to a 3D-stacked memory. We refer to this design as \codebold{FPGA}.
%

In-memory accelerators focus on the inference phase~\cite{tetris:asplos:2017, graphpim:hpca:2017, tom:isca:2016, chameleon:micro:2016, pim:nda:hpca:2015, pim:graph:isca:2015, pim:drama:cal:2015, pim:sparse:hpec:2013, neuflow:cvprw:2011, machine:neurocomputing:2017} or the training phase of a restricted set of ML algorithms such as CNNs~\cite{neurocube:isca:2016}.
As there is no previous in-memory accelerator that supports the training phase of different types of ML algorithms, we compare \hpim~with a design that uses the general-purpose ALUs similar to those in prior work ~\cite{tabla:hpca:2016, cosmic:Micro:2017} on the logic die of the 3D-stacked memory. We refer to this design as \codebold{PIM-GU}.
To understand the limitations of previous in-memory accelerator, we compare the results to an ideal but impractical platform, \codebold{PIM-GU-Unlimited}, with enough general-purpose ALUs on the logic die of the 3D-stacked memory to fully utilize the available bandwidth. 
\codebold{PIM-CE} evaluates an \hpim~which only benefits from the compute engines on the logic die. 
We use this comparison point to show the importance of split execution. Moreover, this comparison point shows the effectiveness of heterogeneous compute engines as compared to general-purpose ALUs.

Table~\ref{tab:syn} shows the specifications of \hpim~accelerators and the ALU of prior work~\cite{tabla:hpca:2016, cosmic:Micro:2017}.
The table shows that we can only include \xx{32} ALUs on the logic die, as the area of one ALU is 1.2~$mm^{2}$ and the available area in a single vault is \code{1.5\,mm\textsuperscript{2}}. Consequently, \codebold{PIM-GU} exploits 80~GB/s of the available bandwidth. 
\\
\begin{table}[h]
    \centering
   \caption{Area, power, and latency of the compute engines of \hpim~and a general-purpose ALU in a \code{45-nm} technology node.} 
    \includegraphics[width=0.35\textwidth]{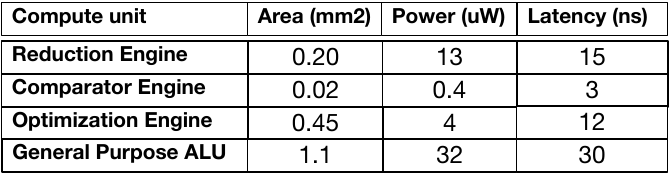}
    \label{tab:syn}
\end{table}
\subsection{Experimental Results}
\label{sec:eval}

\begin{figure*}[h]
    \centering
    \includegraphics[width=1\textwidth]{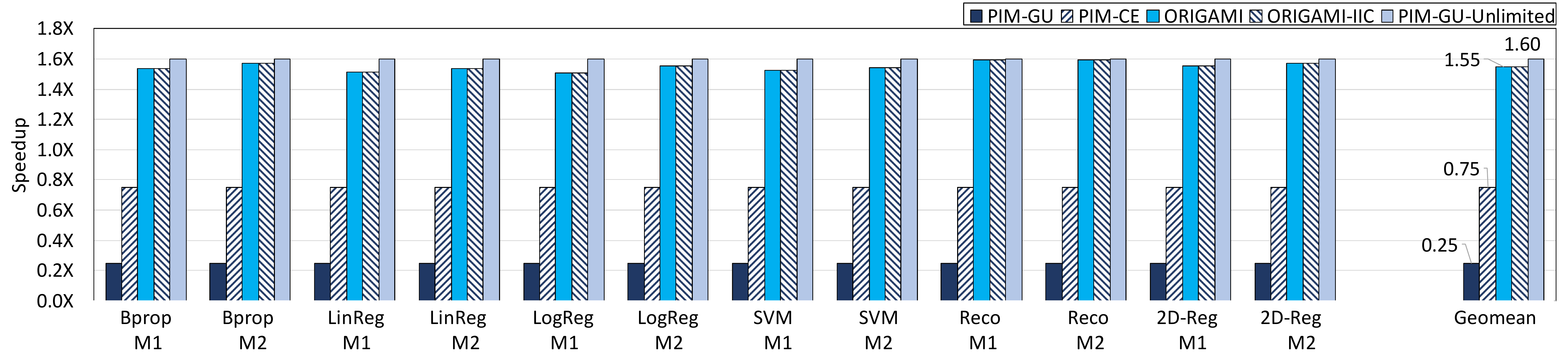}
    \caption{Speedup of the competing compute platforms over \codebold{FPGA}.}
    \label{fig:speedup}
\end{figure*}
\begin{figure*}[h]
    \centering
    \includegraphics[width=1\textwidth]{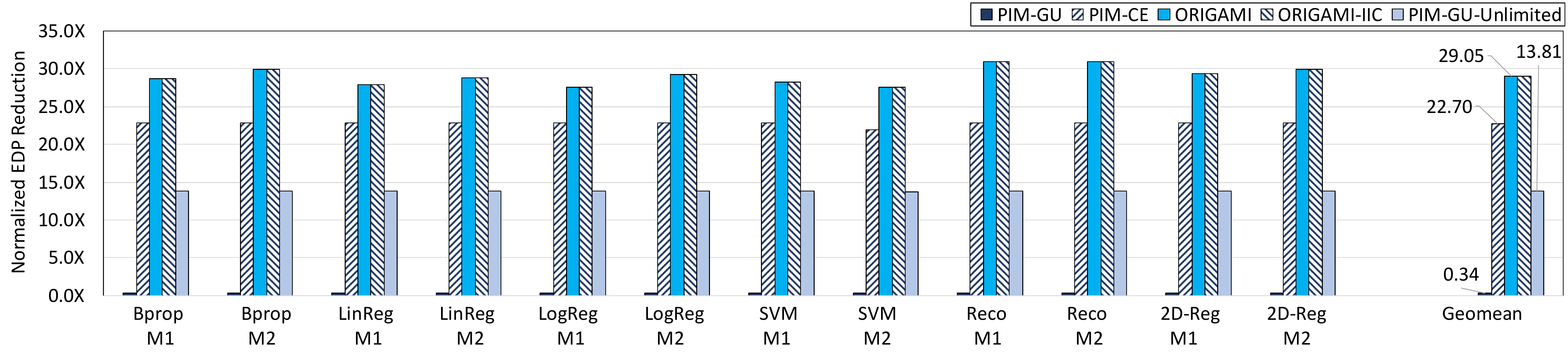}
    \caption{EDP reduction of the competing platforms, normalized to \codebold{FPGA}.}
    \label{fig:energy}
\end{figure*}
\niparagraph{Performance Analysis.}
To evaluate the effect of \hpim~on accelerating the training phase of ML algorithms, we measure the execution time across the evaluated benchmarks.
Figure~\ref{fig:speedup} shows the speedup (higher is better) of different platforms with respect to \codebold{FPGA}.
We make four key observations.

%
First, \codebold{\hpim}~outperforms \codebold{FPGA} in terms of execution time, by \xxs{1.55$\times$} on average (up to \xxs{1.6$\times$}).
\codebold{\hpim}~exploits all the available bandwidth, while \codebold{FPGA} only captures the external bandwidth. 
%
Second, The speedup of \codebold{\hpim}~is within \xx{$\approx$1\%} of \codebold{PIM-GU-Unlimited}. 
The reason for this level of speedup is the effectiveness of \codebold{\hpim}~in capturing all the available memory bandwidth. 
\codebold{\hpim}~maximally utilizes the memory bandwidth by judiciously distributing the computations between the FPGA and the accelerators on the logic die of the 3D-stacked memory. 

Third, \codebold{\hpim}~offers the same speedup as \codebold{\hpim-IIC}, which shows how well our partitioning algorithm minimizes the inter-platform communications. Our evaluations show that the bandwidth overhead in \codebold{\hpim}~is less than \xxs{0.001\%} and \codebold{\hpim}~effectively hides the delay overhead.
%
%
Fourth, \codebold{PIM-CE} outperforms \codebold{PIM-GU} by \xxs{2.9$\times$}, which shows the effectiveness of the heterogeneous compute engines as compared to general-purpose units. 
Moreover, \hpim~combines heterogeneous compute engines with split execution to capture the whole available bandwidth, and hence, outperforms~\codebold{PIM-CE} by \xxs{2.1$\times$}.
%

%
%

\niparagraph{Energy Analysis.}
We measure energy-delay product (EDP) of all benchmarks on all compute platforms.
Figure~\ref{fig:energy} shows the normalized EDP reduction (higher is better) of the compute platforms, normalized to \codebold{FPGA}.

We make four key observations. 
First, \codebold{\hpim}~outperforms \codebold{FPGA} by \xxs{29$\times$}, on average (up to \xxs{31$\times$}). It is due to two reasons: (1) in \codebold{\hpim}, a portion of data communications is local, as it executes a portion of computations inside the 3D-stacked memory, and (2) \codebold{\hpim}~exploits light-weight compute engines in both in-memory and out-of-the-memory platforms, which consume less energy than the general-purpose ALUs of \codebold{FPGA}. 
%
%
%
Second, EDP of \codebold{\hpim}~is \xxs{86.1$\times$} and \xxs{2.1$\times$} lower than \codebold{PIM-GU}'s and \codebold{PIM-GU-Unlimited}'s, respectively. Although \codebold{PIM-GU} and \codebold{PIM-GU-Unlimited} perform all the communications and computations inside the 3D-stacked memory, their general-purpose ALUs consume more energy than compute engines of \codebold{\hpim}.
Third, \codebold{PIM-CE} outperforms both \codebold{PIM-GU} and \codebold{PIM-GU-Unlimited} by \xxs{67.3$\times$} and \xxs{1.7$\times$}, on average, respectively. This is because \codebold{PIM-CE} exploits heterogeneous compute engines whose energy usage is significantly lower than general-purpose ALUs in \codebold{PIM-GU} and \codebold{PIM-GU-Unlimited}.
Fourth, \codebold{\hpim}~and \codebold{\hpim-IIC}~offer very close EDP due to low inter-platform communications offered by split execution of \codebold{\hpim}.
%

%


\begin{figure}[h]
    \centering
    \includegraphics[width=0.35\textwidth]{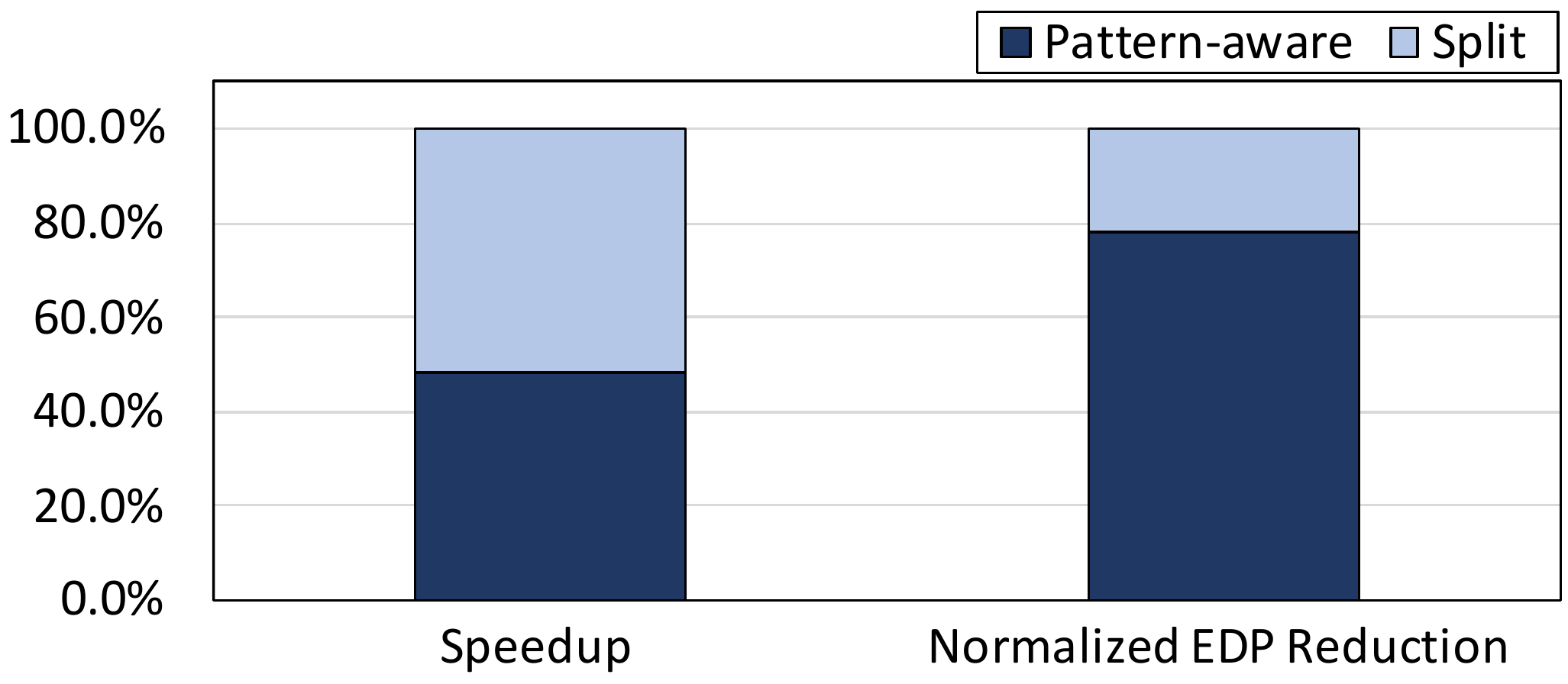}
    \caption{Breakdown of speedup and EDP reduction between pattern-aware execution and split execution of \hpim.}
    \label{fig:breakdown}
\end{figure}
%
\begin{figure*}[h]
    \centering
    \includegraphics[width=1\textwidth]{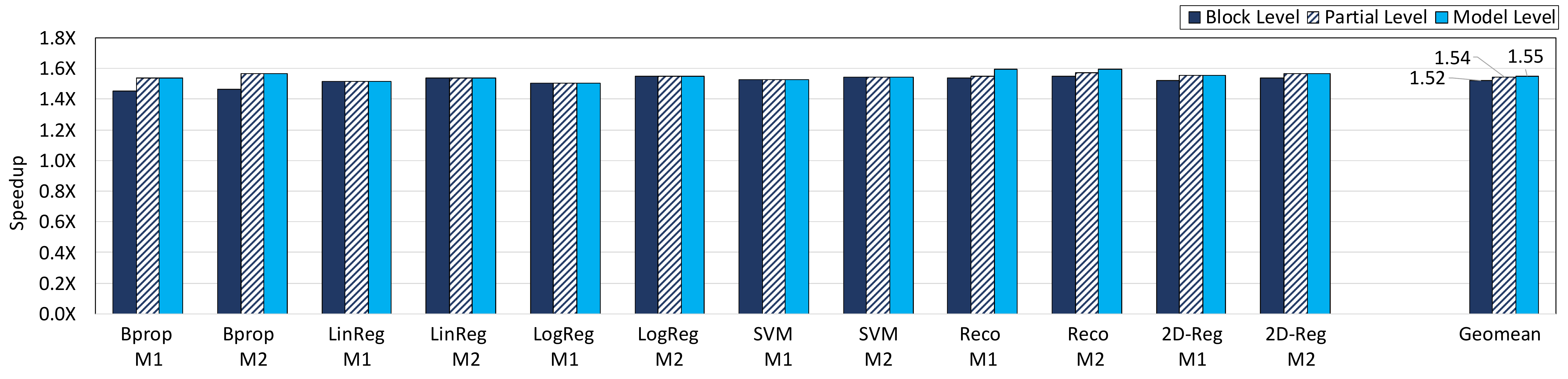}
    \caption{Speedup sensitivity to the three parallelism types.}
    \label{fig:speedup-parallel}
\end{figure*}
\begin{figure*}[h]
    \centering
    \includegraphics[width=1\textwidth]{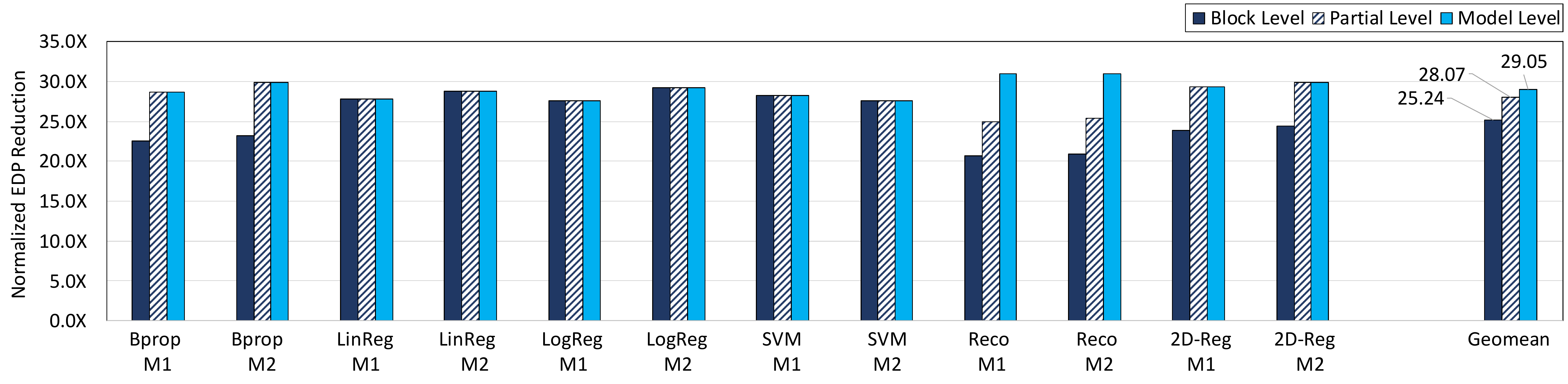}
    \caption{EDP sensitivity to the three parallelism types.}
    \label{fig:energy-parallel}
\end{figure*}
\niparagraph{Sources of Benefit.}
\codebold{\hpim} exploits two execution techniques, \emph{pattern-aware execution} and \emph{split execution}, to effectively run different kinds of ML algorithms. To shed light on the importance of these two techniques, we compare \codebold{\hpim} against \codebold{PIM-GU}. As compared to \codebold{PIM-GU}, \codebold{\hpim} offers two advantages: (1) heterogeneous compute engines instead of general-purpose ALUs, and (2) split execution over the compute engines and the out-of-the-memory platform.
Figure~\ref{fig:breakdown} shows the contribution of each technique to the speedup and EDP reduction of \codebold{\hpim} as compared to \codebold{PIM-GU}. The figure shows that heterogeneous compute engines are responsible for 48\% of the speedup and 78\% of the EDP reduction. Likewise, split execution is responsible for 52\% of the speedup and 22\% of the EDP reduction. These results clearly show that both techniques are necessary for the success of \codebold{\hpim}.
%

%
\niparagraph{Sensitivity to Parallelism Types.}
To better illustrate the source of benefit in \codebold{\hpim}, Figures~\ref{fig:speedup-parallel} and ~\ref{fig:energy-parallel} show the effect of different parallelism types (i.e., \emph{block\_{level}}, \emph{partial\_{level}}, and \emph{model\_{level}}) on the speedup and EDP across the evaluated benchmarks.
The first bar, \codebold{block\_{level}}, shows the results when only \emph{block\_{level}} parallelism is enabled.
The second bar, \codebold{partial\_{level}}, shows the speedup and EDP when both \emph{block\_{level}} and \emph{partial\_{level}} parallelism types are enabled.
Finally, the last bar, \codebold{model\_{level}}, illustrates the results when all three parallelism types (the default mode in \codebold{\hpim}) are enabled.
We make three observations.

First, by enabling \emph{block\_{level}} parallelism, \codebold{\hpim}~outperforms \fpga in terms of speedup and EDP, by \xxs{1.52$\times$} and \xxs{25.24$\times$}, on average, respectively.
It is due to the fact that all benchmarks benefit from the \emph{block\_{level}} parallelism, thus by leveraging \emph{block\_{level}} parallelism, \codebold{\hpim}~splits the execution over both compute engines in 3D-stacked memory and the out-of-the-memory FPGA platform.
Out of twelve benchmarks, six benchmarks (\bench{LinReg (M1), LinReg (M2), LogReg (M1), LogReg (M2), SVM (M1), and SVM (M2)} have only \emph{block\_{level}} parallelism. 
Thus, these benchmarks see no speedup or improvement in EDP by enabling \emph{partial\_{level}} and \emph{model\_{level}} parallelism types.

Second, by enabling both~\codebold{block\_{level}}~and~\codebold{partial\_{level}}, on average, \codebold{\hpim}~achieves \xxs{1.56$\times$} and \xxs{28.10$\times$} improvement in execution time and EDP over \codebold{\fpga}.
\bench{BProp}, \bench{Reco}, and \bench{2D-Reg} achieve higher speedup and lower EDP over \codebold{block\_{level}}.
As an example, \codebold{partial\_{level}} improves the speedup and EDP of \bench{BProp} by \xxs{$\approx$10\%} and \xxs{26.75\%}, respectively, over \codebold{block\_{level}}.
%
%
%
%

Third, by enabling all three parallelism types, \codebold{\hpim} improves speedup and EDP of \bench{Reco} by \xxs{1.6$\times$} and \xxs{31.0$\times$}, on average, respectively, over \codebold{\fpga}.
Other benchmarks do not have \codebold{model\_{level}} parallelism and achieve no improvements by enabling \codebold{model\_{level}}.
\bench{Reco} has two independent models and achieves the highest speedup and the lowest EDP by \codebold{model\_{level}}.
\codebold{model\_{level}} improves the speedup and EDP reduction of \bench{Reco} as compared to other parallelism types by \xxs{6\%} and \xxs{1.50$\times$}, on average, respectively.
Although \bench{BProp} includes three models, they are not independent, thus, \bench{BProp} does not have \codebold{model\_{level}} parallelism. 
%
%
%
These results asserts the importance of exploring various types of parallelism to fully benefit from \codebold{\hpim}. 
\section{Related Work}
\label{sec:related}
Our proposal, \hpim, is fundamentally different from prior work in the following directions:
(1) \hpim~extracts compute patterns of ML algorithms and translates them into heterogeneous compute engines on the logic die of a 3D-stacked memory,
(2) \hpim~splits execution of ML algorithms over the heterogeneous compute engines and an out-of-the-memory compute platform to utilize all the available bandwidth, and (3) \hpim~exploits an optimization algorithm to split the computation of ML algorithms between two platforms in a load-balanced manner and with minimum inter-platform communications to maximize resource utilization.  

There has been a wealth of architectures for in-memory accelerators that integrate logic and memory onto a single die to enable higher memory bandwidth and lower access energy~\cite{tetris:asplos:2017,pipelayer:hpca:2017,tom:isca:2016,isaac:archnews:2016,neurocube:isca:2016,chameleon:micro:2016,prime:archnews:2016,pim:nda:hpca:2015,pim:drama:cal:2015,pim:sparse:hpec:2013,neuflow:cvprw:2011, bitfusion:isca:2018, snapea:isca:2018, ganax:isca:2018, saberlda:asplos:2017, eyeriss:isca:2016, cnvlutin:rchnews:2016, shidiannao:isca:2015, stripes:micro:2016, scnn:isca:2017, eie:isca:2016, cambrion:micro:2016, dadiannao:micro:2014, pudiannao:asplos:2015, FPGADeep:FPGA:2015, dnnweaver:micro:2016,tabla:hpca:2016, CHiMPS:FPGA:2008, large:journal:2011, tensorflow, convGPU:2017, Oh:2004, Guzhva:2009, tpu:isca:2017, CMP-PIM_DAC118, PIM-training_Micro18}.
Most of these in-memory architectures accelerate the inference phase of ML algorithms, some of in-memory accelerators, such as Neurocube~\cite{neurocube:isca:2016} and Proger PIM~\cite{PIM-training_Micro18}, accelerate both the training and inference phases, only target CNNs and do not work for other ML algorithms.

Prior work exploits ASIC~\cite{bitfusion:isca:2018, snapea:isca:2018, ganax:isca:2018, eyeriss:isca:2016, cnvlutin:rchnews:2016, shidiannao:isca:2015, stripes:micro:2016, scnn:isca:2017, eie:isca:2016, cambrion:micro:2016, dadiannao:micro:2014, pudiannao:asplos:2015, tpu:isca:2017}, GPU~\cite{convGPU:2017, Oh:2004, Guzhva:2009, saberlda:asplos:2017}, FPGA~\cite{FPGADeep:FPGA:2015, dnnweaver:micro:2016,tabla:hpca:2016, CHiMPS:FPGA:2008, large:journal:2011}, and multi-computing-node~\cite{tensorflow, tpu:isca:2017} platforms to accelerate ML algorithms. While effective, these techniques do not benefit from in-memory processing.
Some prior work used split execution to accelerate ML algorithms.
Shen, et al. ~\cite{resourcePartitioning:ISCA:2017} partitioned FPGA resources to process different subsets of convolutional layers of CNNs. 
Scalpel~\cite{scalpel:ISCA:2017} customizes DNN pruning over SIMD-aware weight pruning and node pruning.
Park, et al.~\cite{cosmic:Micro:2017} distribute only the optimization part of the training phase of different ML algorithms over FPGA and \emph{n} ASIC units. Consequently, their technique does not offer load balancing.
Scaledeep~\cite{scaledeep:ISCA:2017} uses heterogeneous processing tiles that are customized for compute-intensive and memory-intensive parts of training DNNs.
Proger PIM~\cite{PIM-training_Micro18} uses CPU, fixed-function PIM and a programmable PIM unit for training different CNN models. 
These techniques are fundamentally different from our proposed technique since none of them is in-memory processing and they do not offer all the necessary features needed for an efficient in-memory processing.
%
\section{Conclusion}
\label{sec:concl}

During the training phase, ML algorithms process large amounts of data, iteratively, which consumes significant bandwidth and energy.
Although in-memory accelerators provide high memory bandwidth and consume less energy, they suffer from lack of generality or efficiency. 
We propose \hpim, a holistic approach that exploits heterogeneous compute engines on the logic die to efficiently cover a wide range of ML algorithms and splits the execution of ML algorithms over the in-memory compute engines and an out-of-memory compute platform to use all the available bandwidth. 
The evaluation results show that \hpim~outperforms the best-performing prior work, in terms of performance and energy-delay product (EDP), by up to \xxs{1.6$\times$} and \xxs{31$\times$}, respectively. \hpim~also improves average performance and energy efficiency by \xxs{1.5$\times$} and \xxs{21$\times$}, respectively. 
%

\bibliographystyle{ieeetr}
\bibliography{paper}

\end{document}